\documentclass[letterpaper]{article} 
\usepackage{aaai25}  
\usepackage{times}  
\usepackage{helvet}  
\usepackage{courier}  
\usepackage[hyphens]{url}  
\usepackage{graphicx} 
\def\UrlFont{\rm}  
\usepackage{natbib}  
\usepackage{caption} 
\usepackage{algorithm}
\usepackage{algorithmic}
\usepackage{booktabs}
\usepackage{subfigure}
\usepackage{subcaption}
\usepackage{newfloat}
\usepackage{listings}
\DeclareCaptionStyle{ruled}{labelfont=normalfont,labelsep=colon,strut=off} 
\floatstyle{ruled}
\newfloat{listing}{tb}{lst}{}
\floatname{listing}{Listing}
\title{Evolutionary Large Language Model for Automated Feature Transformation}
\author {
    Nanxu~Gong\textsuperscript{\rm 1},
    Chandan~K Reddy\textsuperscript{\rm 2},
    Wangyang~Ying\textsuperscript{\rm 1},
    Haifeng~Chen\textsuperscript{\rm 3},
    Yanjie~Fu\textsuperscript{\rm 1}\thanks{Corresponding author.}
}
\affiliations {
    \textsuperscript{\rm 1}Arizona State University, Tempe, USA\\
    \textsuperscript{\rm 2}Virginia Tech, Arlington, USA\\
    \textsuperscript{\rm 3}NEC Laboratories America, Princeton, USA\\
    nanxugong@outlook.com,
    reddy@cs.vt.edu,
    wangyang.ying@asu.edu,
    haifeng@nec-labs.com,
    yanjie.fu@asu.edu
}

\usepackage{bibentry}

\begin{document}

\maketitle

\begin{abstract}
Feature transformation aims to reconstruct the feature space of raw features to enhance the performance of downstream models. However, the exponential growth in the combinations of features and operations poses a challenge, making it difficult for existing methods to efficiently explore a wide space. Additionally, their optimization is solely driven by the accuracy of downstream models in specific domains, neglecting the acquisition of general feature knowledge. To fill this research gap, we propose an evolutionary LLM framework for automated feature transformation. This framework consists of two parts: 1) constructing a multi-population database through an RL data collector while utilizing evolutionary algorithm strategies for database maintenance, and 2) utilizing the ability of Large Language Model (LLM) in sequence understanding, we employ few-shot prompts to guide LLM in generating superior samples based on feature transformation sequence distinction. Leveraging the multi-population database initially provides a wide search scope to discover excellent populations. Through culling and evolution, high-quality populations are given greater opportunities, thereby furthering the pursuit of optimal individuals. By integrating LLMs with evolutionary algorithms, we achieve efficient exploration within a vast space, while harnessing feature knowledge to propel optimization, thus realizing a more adaptable search paradigm. Finally, we empirically demonstrate the effectiveness and generality of our proposed method. The code is available at \url{https://github.com/NanxuGong/ELLM-FT}.
\end{abstract}
\section{Introduction}
In many real-world applications, ML models struggle to fight complex and imperfect data (e.g., bias, outliers, noises).
The quality of data, as a fundamental element in machine learning (ML), plays a significant role in the predictive performance of ML.  To alleviate this issue, feature transformation is proposed to reconstruct an optimized feature space based on original features and mathematical operations (e.g., +, -, *, /, sqrt). In industrial practices, traditional feature transformations are typically labor intensive, time costly, and lack generalization. Therefore, we focus on the task of Automated Feature Transformation (AFT) \cite{wang2024reinforcement,wang2022group,kanter2015deep} that aims to reconstruct a discriminative feature space through an automatic model.

There are two main challenges in solving AFT: 1) efficient search in massive discrete space; and 2) teaming between general feature knowledge and task-specific feature knowledge. 
First, a feature space comprises exponentially growing possibilities of combinations of features and operations as candidate feature transformations, resulting in an immensely large search space. Efficient search in massive discrete space aims to answer: \textit{How can we improve the efficiency of identifying the optimal search path given a large feature combination space?} 
Second, we need knowledge to steer the optimal search path. A widely used idea is to exploit task-specific feature knowledge, defined as predictive accuracy feedback of a transformed feature set on a downstream ML model. This strategy ignores general feature knowledge from Artificial General Intelligence (AGI) like ChatGPT and other LLMs. Teaming between general feature knowledge and task-specific knowledge aims to answer: \textit{How can we steer the optimal search path by leveraging both task-specific feature knowledge and LLM-like AGI?}

Prior literature can partially address the challenges. These methods can be divided into three categories:
(1) \textit{expansion-reduction methods} \cite{kanter2015deep,khurana2016cognito}. These methods employ operators and randomly combine features to generate new feature samples, expanding the feature space. Subsequently, useful features are further filtered through feature selection. However, such methods rely on stochasticity, and lack optimization trajectories. 
(2) \textit{evolution-evaluation methods} \cite{wang2022group,khurana2018feature}. These methods, rooted in reinforcement learning or evolutionary algorithms, amalgamate features and operator sets within a unified learning framework. They iteratively generate improved individuals until the model reaches the maximum iteration threshold. Such methods generally harbor explicit objectives. However, achieving maximal rewards over the long term is challenging. In open environments, they tend to fall into local optima. 
(3) \textit{Neural Architecture Search (NAS)-based methods} \cite{chen2019neural,zhu2022difer}. These methods are constructed upon NAS, which was proposed to search the optimal network architectures. Given that the objectives of AFT align closely with those of NAS, related methods have also been applied in this domain. However, such methods struggle to model the expansive feature transformation space while exhibiting diminished efficiency. Existing studies show limitations on jointly addressing efficiency and task-specific and general feature knowledge teaming in feature transformation. As a result, we need a novel perspective to derive a new formulation for AFT.

\textbf{Our insights: an evolutionary LLM generation perspective.} We formulate the feature transformation as a sequential generation task. We regard a transformed feature set as a token sequence comprising feature ID symbols and operators.
The emerging LLM (e.g., ChatGPT) has shown its few-shot and in-context learning ability to optimize and generate through seeing demonstrations. Additionally, LLMs can robustly optimize the feature space using both general feature knowledge and task-specific knowledge. Our first insight is to leverage LLMs as a feature transformation generator by demonstrating sample feature transformation operation sequence and corresponding priority to LLMs, so that LLMs can progressively learn complex feature knowledge, capture feature-feature interactions, and discern optimization directions.
Our second insight is to combine LLM with Evolutionary Algorithms (EA) to obtain an evolutionary LLM. In feature transformation contexts, EA can serve as a decision science model to decide the order, quality, and diversity of few-shot demonstrations, strengthen few-shot learning, and alleviate the hallucination of LLMs.


\textbf{Summary of Proposed Approach.} In this paper, we propose a novel \textbf{E}volutionary \textbf{L}arge \textbf{L}anguage \textbf{M}odel framework for automated \textbf{F}eature \textbf{T}ransformation (\textbf{ELLM-FT}). The framework has two goals: 1) LLM as a feature transformation operation sequence generator; and 2) teaming LLMs with EA for better few-shot demonstrations to identify the optimal search direction. 
To achieve these goals, we first leverage a reinforcement learning data collector to construct a multi-population dataset. Within the database, each population evolves independently. We progressively eliminate subpar individuals while adding generated high-quality ones, and likewise eliminate inferior populations, granting superior populations more opportunities for evolution.  We keep the diversity of the database from a multi-population perspective while continuously enhancing the quality of samples within the population. We craft meticulously designed prompts for the pre-trained LLM. By leveraging the feature knowledge of few-shot samples, we guide the LLM to uncover optimization directions. Throughout the iterative process, we continuously update the database to enhance the performance of the LLM with diverse and high-quality samples. By employing \textit{Llama-2-13B-chat-hf} as the backbone of LLM, ELLM-FT demonstrates competitive performance in diverse datasets.

\textbf{Our contributions:} 1) We formulate feature transformation as LLM generative few-shot learning to show that LLM can learn complex feature interaction knowledge and generate improved feature spaces by observing demonstrations of feature transformation operation sequences; 2) We develop an evolutionary LLM framework to show that teaming LLM with EA can enable collaboration between task-specific knowledge and general knowledge, improve demonstration quality, and identify a better optimization direction; 3) We compare ELLM-FT with 7 widely used methods across 12 datasets. Our experimental results demonstrate the effectiveness and robustness of ELLM-FT.


\section{Problem Statement}

Feature transformation utilizes mathematical operations to cross original features, in order to generate new features and reconstruct a better feature space. Learning feature transformation usually requires general feature knowledge and task-specific knowledge. Pre-trained on large-scale datasets, LLMs can leverage general knowledge to understand features and operators at the token level. Maintaining a diverse and high-quality database through evolutionary algorithms provides LLMs with task-specific knowledge, facilitating the generation of novel and superior samples. Thus, the task of feature transformation can be formulated in the evolutionary LLM perspective.

\begin{figure}[t]
    \centering
    \includegraphics[width=\linewidth]{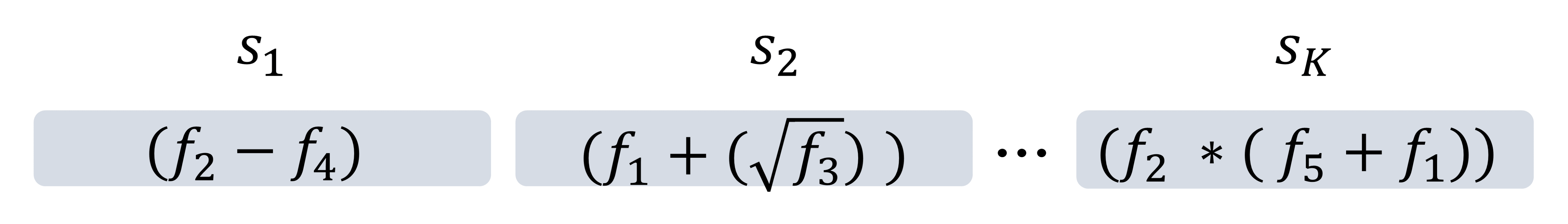}
    \caption{An example of feature transformation sequence. $s_i$ denotes the feature sequence.}
    \label{fig:sequence}
    \vspace{-0.5cm}
\end{figure}

Formally, we define the mathematical operation set $\mathcal{O}$ (e.g., ``log", ``sin", ``plus") and the original feature set $X = [f_1, f_2, ..., f_n]$
. A transformed feature set (e.g., $f1, f2+f1, f3/f2, \sqrt{f4}$) is regarded as a token sequence of feature IDs and operators (\textbf{Figure \ref{fig:sequence}}). The feature transformation task is then seen as a sequential generation task. We leverage LLM to generate feature transformation sequences by demonstrating few-shot examples. Specifically, we firstly construct a database of feature sets and corresponding task performance, denoted by  $\mathcal{D} = \{X,y\}$. We denote a pre-trained LLM with the parameter $\theta$ by $\mathcal{F}_\theta$. We feed a designed prompt $p$, including instructions and few-shot examples, denoted by$X \sim \mathcal{F}_\theta(\cdot \vert p)$, to LLM in order to generate feature transformation sequences. 
Our goal is to maximize the utility score  $Score_{\phi}(X,t)$, where $\phi$ denotes the specific problem and $t$ denotes the target. 
Finally, the ELLM-FT problem is given by: $argmax_{X \sim \mathcal{X}_\theta} Score_{\phi}(X,t)$, where $\mathcal{X}_\theta$ is the set of sequences generated by $\mathcal{F}_\theta$.

\section{Evolutionary LLM For Generative Feature Transformation}

\begin{figure*}[t]
    \centering
    \includegraphics[width=\linewidth]{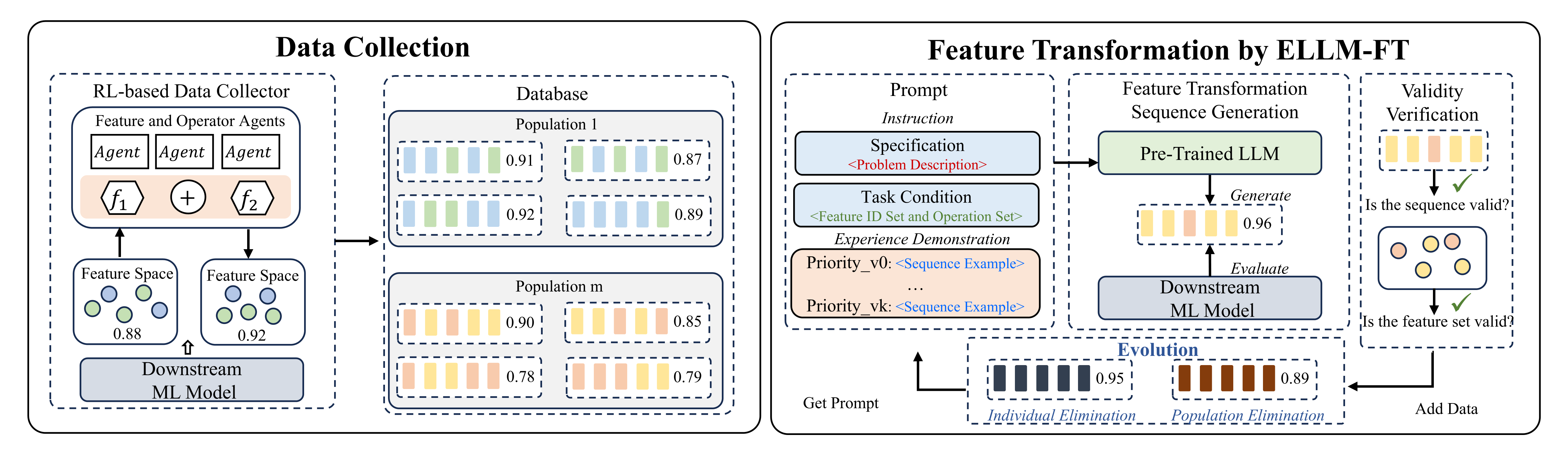}
    \caption{Framework overview. Firstly, we utilize the RL data collector to construct the database, Then, we leverage pre-trained LLM to iteratively generate new feature transformation sequences while simultaneously updating the database.}
    \label{fig:framework}
    \vspace{-0.4cm}
\end{figure*}

\subsection{Overview of the ELLM-FT Framework}
Figure \ref{fig:framework} shows our method includes two steps:
1) reinforcement multi-population training database construction; 
2) LLM-based feature transformation sequence generation.
Step 1 is to develop a reinforcement training data collector. 
This collector has two feature selection agents and an operation selection agent. In each reinforcement step, feature crossing-generated features are added to the previous feature set to create a transformed feature set. Each reinforcement episode includes multiple steps and thus collects multiple feature transformation sequences as a population. 
Step 2 is to iteratively guide the LLM through few-shot prompts to discover optimization directions from the differences among ranked samples, thereby generating better samples. Step 2 includes four components: sequential promoting design,  feature transformation sequence generation, verification and evaluation, and database updating. 

\subsection{Reinforcement Multi-Population Training Database Construction} 
\textbf{Why Constructing A Multi-Population Database?} 
We propose to see LLM as a feature transformation sequence generator. We want LLM to learn from few shot examples of historical feature transformation sequences via smart prompting. 
Therefore, it is needed to develop a database of feature transformation sequences training samples along with corresponding downstream task performance. 
The multi-population concept is to diversify the training database so its distribution can robustly cover most areas of the true distribution of the feature transformation sequence space. 

\noindent\textbf{Reinforcement Transformation-Performance Data Collection.}  It is traditionally challenging to collect diverse and high-quality training data of feature transformations.
Manual data collection is time-costly.
Inspired by~\cite{wang2022group, xiao2023traceable}, we develop a reinforcement training data collector to explore and collect various feature transformations and corresponding downstream task performance. 
The reinforcement data collector includes: 
1) \textit{Three agents:} a head feature agent, an operation agent, and a tail feature agent, respectively denoted by $\alpha_h, \alpha_o$, and $\alpha_t$. 
2) \textit{Actions:} two feature agents select features from the feature set $X$ and the operation agent selects operators from the operator set $\mathcal{O}$. At the $t$-th iteration, the three agents collaborate to generate a new feature, denoted by $f_t = \alpha_h(t) \oplus\alpha_o(t) \oplus\alpha_t(t)$. 
3) \textit{State:} to help agents understand the current feature space, we extract the first-order and second-order descriptive statistics of the feature set as a state representation vector, denoted by $S_t = Rep(X_t)$. Specifically, given a data matrix, we first compute the 7 descriptive statistics (i.e. count, standard deviation, minimum, maximum, first, second, and third quartile) of each column, then compute the same 7 descriptive statistics of the column-wise descriptive statistics row by row. The row-wise statistics of column-wise statistics of the data matrix form a representation vector of 49 variables.
4) \textit{Reward:} we define the reward as the improvement of downstream model accuracy, denoted by: $R(t) = y_{t} - y_{t-1}$. 
Finally, we minimize the mean squared error of the Bellman Equation to optimize the procedure. Algorithm \ref{alg:rl} shows in the t-th step, a transformed feature set is created, in the i-th episode, a group of transformed feature sets, also known as, a population of feature transformation sequences, is created. 
Specifically,  during the $t$-th step, with the previous feature space denoted as $X_{t-1} = [f_1, f_2, ...,f_{t-1}]$, three agents collaborate to select two features and one operator to generate a new feature $f_{t}$. By adding $f_{t}$ into $X_{t-1}$, we obtain a new sample \{$X_t, y_t$\}, where $X_t$ denotes the new transformed feature set and $y_t$ is the corresponding downstream ML model accuracy. 
The state is updated by $Rep(X_t)$, where $Rep$ is the representative function to extract the descriptive statistics of $X_t$.

\noindent\textbf{Creating Populations Per Episode.} 
To create a multi-population database, we regard the samples collected in one episode (indexed by $t$) as a population. 
Be sure to notice that in one population (i.e., episode), the collected feature transformation sequences (also known as transformed feature sets) can be seen as a Markov decision process (MDP) with current-previous dependency. In other words, in this population, a feature transformation token sequence is built yet improved based on another MDP-dependent sequence. 
So, LLM can easily identify shared patterns and contrast differences to better learn optimization directions.   

\begin{algorithm}[h]
    \caption{RL-based data collection}

    \label{alg:rl}
    \renewcommand{\algorithmicrequire}{\textbf{Input:}}
    \renewcommand{\algorithmicensure}{\textbf{Output:}}
    
    \begin{algorithmic}[1]
        \REQUIRE The original feature set $X$, downstream ML model $\phi$, representative function $Rep$, episode number $M$, step number $T$.  
        \ENSURE Feature transformation dataset $\tau$.   
        
        \FOR{$i =$ 1 to $M$} 
            \STATE Initialize the state $s_0$
            \FOR{$t =$ 1 to $T$}
            \STATE $f_t \leftarrow \alpha_h(t) \oplus \alpha_o(t) \oplus \alpha_t(t)$
            \STATE $X_t \leftarrow X_{t-1} + f_t$
            \STATE $y_t \leftarrow \phi (X_t)$
            \STATE $s_t \leftarrow Rep (X_t)$
            \STATE $R_t \leftarrow y_{t} - y_{t-1}$
            \STATE $\tau \leftarrow \tau +\{X_t, y_t\}$
            \STATE Updated by Bellman Equation
            \ENDFOR
        \ENDFOR
        \RETURN $\tau$.
    \end{algorithmic}
\end{algorithm}

\begin{figure*}[t]
    \centering
    \includegraphics[width=0.75\linewidth]{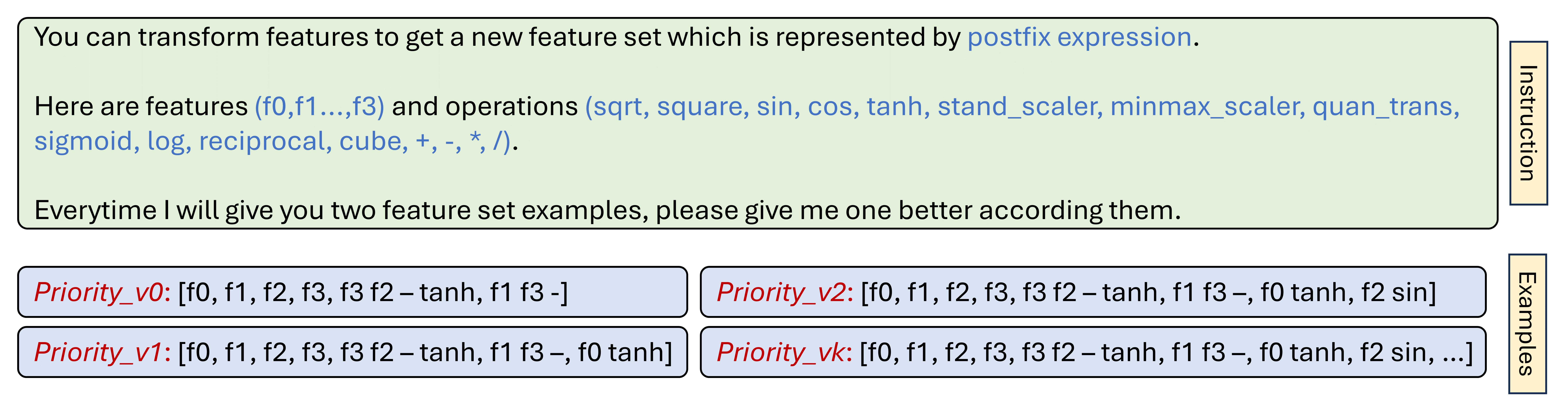}
    \caption{An example of our prompt consisting of the instruction and few-shot feature transformation operation sequence samples.}

    \label{fig:prompt}
    \vspace{-0.5cm}
\end{figure*}

\noindent\textbf{Representing Feature Transformation Sequences with Postfix Expressions.}
A sequence of feature transformations can be presented by infix, prefix, or postfix. 
Postfix has a number of advantages over other expressions. Compared with infix, a token sequence can be expressed without parentheses via postfix.  Second, compared with prefix,  postfix is easier to implement evaluation. 
Therefore, we convert feature transformation sequences into postfix expressions. For example, the sequence $\{f_0, (f_1 + f_2), ((f_0 + f_2) * f_3)\}$ can be represented by $\{f_0, f_1 f_2 +, f_0 f_2 + f_3 *\}$. 

\subsection{LLM-based Feature Transformation Operation Sequence Generation}

\textbf{Why Generating Feature Transformation Sequences via LLM?} 
Firstly, LLM is a generic generator, and, thus, can be used to generate feature transformation sequences. 
Secondly, LLM exhibits few-shot learning abilities; in other words, there is to enable LLM to learn feature space knowledge and the optimization direction of sequence generation, by demonstrating a list of feature transformation sequence examples in an instructional prompt. 
Thirdly, by developing an evolutionary prompt, LLM can iteratively self-optimize the generation quality over time. 

\noindent\textbf{Step 1: Sequential Prompting Design.}
To enable LLM to perceive the optimization direction of generating feature transformation sequences, we need a unique sequential prompting design. Existing studies on sequential prompting have proposed two design strategies: 
1) randomly sample $M$ examples from a population~\cite{meyerson2023language}; 
2) sample $M$ top-performing examples from a population and arrange the $M$ examples in terms of performance \cite{romera2024mathematical}. 
However, both strategies have limitations. 
The former is limited by its randomness and thus lacks the articulate optimization direction information for LLM to leverage. The latter always selects top-M samples; in other words, top-performing samples are repeatedly selected, lacking novelty and diversity. 
To fill this gap, we propose a prompt that balances randomness and quality. Figure \ref{fig:prompt} shows an example of our prompt. Our prompt includes two components: instruction and examples. 
The instruction component consists of a task description, an operator set, and original feature IDs. 
The examples component includes few-shot examples of feature transformation operation sequences.
Consider the existence of K populations, during each iteration, we create K prompts corresponding to K populations. 
For the  $k$-th prompt, we randomly select a certain number of feature transformation sequence samples from the $k$-th population. 
Then, we rank these samples in the k-th prompt in terms of their downstream task performance (namely priority\_v0, priority\_v1, ...). 
In this way, LLM can observe a list of progressively improving feature transformation sequences in the k-th prompt,  examine the differences between these sequences, moreover, uncover optimization direction to produce a better feature transformation sequence.
At the end of this iteration, the K prompts corresponding to K populations trigger LLM to generate K feature transformation sequences. 
Here, the randomness of samples is from the random selection of samples in a population. The quality of samples is from the quality of the population, which is maintained by a database updating strategy (refer to Step 4: database updating). 

\noindent\textbf{Step 2: Feature Transformation Operation Sequence Generation by LLM.} 
We regard LLM as a feature transformation sequence generator.  
We utilize \textit{Llama-2-13B-chat-hf} as a pre-trained LLM. 
In each iteration, we first select a population and then select a subset of feature transformation sequences from the selected population. We use the feature transformation sequence subset, ranked by corresponding performance, to create a prompt that includes instructions and examples.  
This prompt is later fed into LLM to guide LLM to generate an improved feature transformation sequence. 
Our insight is to leverage the generic knowledge of LLM, and the domain knowledge of the prompt (task instruction and the ranked list of feature transformation sequences) to inspire LLM to perceive optimization directions and generate a better feature transformation sequence. 

\noindent\textbf{Step 3: Verification and Evaluation.}
During each iteration, after LLM generates a feature transformation sequence triggered by a prompt, we conduct the verification of the LLM-generated feature transformation sequence from three perspectives:
1) Are the tokens of the LLM-generated feature transformation sequence from our predefined operator set and feature ID set? 
2) Is the LLM-generated feature transformation sequence formulated as a postfix expression? 
3) Has the LLM-generated feature transformation sequence never appeared in the database before? 
Based on the three criteria, we identify valid feature transformation sequences and store these sequences to the corresponding populations. 
After verification, we conduct the evaluation of the LLM-generated feature transformation sequence. 
We leverage a fixed downstream ML model to evaluate the performance of the LLM-generated feature transformation sequence and associate the performance score with the corresponding feature transformation sequence.

\begin{table*}[!h]
    \centering
    \caption{Comparison of the proposed method with baselines over 12 datasets, where the best results are in bold.}
    \label{tab:overall}
    \resizebox{0.9\linewidth}{!}{
    \begin{tabular}{ccccccccccccc}
   \toprule\toprule
    Dataset & Soruce & C/R & Samples & Features & RDG & ERG & LDA & AFT & NFS & TTG & GRFG& ELLM-FT \\
    \midrule
    Amazon Employee & Kaggle & C & 32769 & 9 & 0.744 & 0.740 & 0.920 & 0.943 & 0.935 & 0.806 & \textbf{0.946} & \textbf{0.946}\\
    SVMGuide3 & LibSVM & C & 1243 & 21 & 0.789 & 0.764 & 0.732 & 0.829 & 0.831 & 0.804 & 0.850 & \textbf{0.856}\\
    German Credit & UCIrvine & C & 1001 & 24 & 0.695 & 0.661 & 0.627 & 0.751 & 0.765 & 0.731 & 0.772 & \textbf{0.775}\\
    Messidor Features & UCIrvine & C & 1150 & 19 & 0.673 & 0.635 & 0.580 & 0.678 & 0.746 & 0.726 & 0.757 & \textbf{0.760}\\
    SpamBase & UCIrvine & C & 4601 & 57 & 0.951 & 0.931 & 0.908 & 0.951 & 0.955 & \textbf{0.961} & 0.958 & 0.957\\
    Ionosphere & UCIrvine & C & 351 & 34 & 0.919 & 0.926 & 0.730 & 0.827 & 0.949 & 0.938 & 0.960 & \textbf{0.963}\\
    \midrule
    Openml\_586 & OpenML & R & 1000 & 25 & 0.595 & 0.546 & 0.472 & 0.687 & 0.748 & 0.704 & 0.783 & \textbf{0.801}\\
    Openml\_589 & OpenML & R & 1000 & 25 & 0.638 & 0.560 & 0.331 & 0.672 & 0.711 & 0.682 & 0.753 & \textbf{0.781}\\
    Openml\_607 & OpenML & R & 1000 & 50 & 0.579 & 0.406 & 0.376 & 0.658 & 0.675 & 0.639 & 0.680 & \textbf{0.793}\\
    Openml\_616 & OpenML & R & 500 & 50 & 0.448 & 0.372 & 0.385 & 0.585 & 0.593 & 0.559 & 0.603 & \textbf{0.739}\\
    Openml\_618 & OpenML & R & 1000 & 50 & 0.415 & 0.427 & 0.372 & 0.665 & 0.640 & 0.587 & 0.672 & \textbf{0.778}\\
    Openml\_620 & OpenML & R & 1000 & 25 & 0.575 & 0.584 & 0.425 & 0.663 & 0.698 & 0.656 & 0.714 & \textbf{0.725}\\
    
    \bottomrule\bottomrule
    \end{tabular}}
    \vspace{-0.4cm}
\end{table*}

\noindent\textbf{Step 4: Database Updating.} 
The quality of samples in the database influences the quality of a prompt, thus, affecting the performance of LLM-generated feature transformation sequences. 
After new samples are added to the database, we propose a two-step database updating strategy: 1) low-quality individual elimination and 2) low-quality population elimination. 
Specifically, to conduct low-quality individual elimination, we devise a population size threshold  $P$. 
Given a population of the size $T$ and $T>P$, we keep top-$P$ quality feature transformation sequences and eliminate top $T-P$ low-quality feature transformation sequences. 
To conduct low-quality population elimination, we score each population using the best performance of all the feature transformation sequences in this population. We then use the top 50\% high-quality populations to replace the top 50\% low-quality populations. 
The updating strategy on the individual and population levels can ensure the quality of populations is improving over time. 
\section{Experimental Results}

\subsection{Experimental Setup}
\noindent\textbf{Data Descriptions.}
We collected 12 datasets from UCIrvine, LibSVM, Kaggle, and OpenML. We evaluated our method and baseline methods on two major predictive tasks: 1) Classification (C); and 2) Regression (R). Table~\ref{tab:overall} shows the detailed statistics of the data sets.

\noindent\textbf{Evaluation Metrics.}
We adopted Random Forest (RF) as the downstream model. We used the F-1 score to measure the accuracy of classification tasks, and use the 1 - relative absolute error (RAE) to measure the accuracy of regression tasks.  We performed 5-fold stratified cross-validation to reduce random errors in experiments.

\noindent\textbf{Baseline Algorithms.}
We compared our method with 7 widely-used feature generation algorithms: (1) \textbf{RDG}, which randomly generates feature-operation-feature transformations to obtain new features. 
(2) \textbf{ERG}, which applies operation on each feature to expand the feature space, then selects valuable features. 
(3) \textbf{LDA}~\cite{blei2003latent}, which learn new features through matrix factorization. 
(4) \textbf{AFAT} \cite{horn2020autofeat}, which iteratively generate new features and leverage multi-step feature selection to pick useful features. 
(5) \textbf{NFS} \cite{chen2019neural}, which models each feature transformation trajectory and optimizes feature generation processes via reinforcement learning. 
(6) \textbf{TTG} \cite{khurana2018feature}, which regard feature transformation as a graph and search the best feature set over the graph via reinforcement learning. 
(7) \textbf{GRFG} \cite{wang2022group}, which leverages a cascading agent structure and feature group crossing to generate new features.

Aside from comparing with the baseline algorithms,  we developed variants of ELLM-FT to investigate the impacts of each technical component. Specifically,
(i) \textbf{ELLM-FT$^f$} adopts the prompt design of \cite{romera2024mathematical} to select the top-$M$ samples from a population and rank the samples in ascending order in each iteration. 
(ii) \textbf{ELLM-FT$^c$} exploits the prompt design of \cite{meyerson2023language} to randomly select $M$ samples in each iteration. 
(iii) \textbf{ELLM-FT$^r$} randomly collect the datal, insteady of using the reinforcement training data collector. 

\subsection{Overall Comparisons}
This experiment aims to answer: \textit{can our feature transformation method outperform other baseline methods on improving downstream task performance?} 
Table \ref{tab:overall} shows the comparisons of our method and 7 baselines on 12 datasets. 
First, although our method fell behind the baseline accuracy by 0.4\% on the single  ``SpamBase" dataset, our method outperformed the best baselines by an average margin of 2.4 \% in overall. 
The underlying driver is that evolutionary LLM can  self-optimize the prompt over iterations to let LLM perceive the optimization direction of feature transformation sequences. 
Second, an interesting finding is that ELLM-FT is more resistant to noisy datasets (e.g., Openml\_616) than baseline algorithms. This can be explained by the argument that optimizing LLM's sequential generation direction by feature knowledge is more generic than downstream task performance feedback-guided search. 
Third, we observed that ELLM-FT generalizes well and strives a balance performance over both large sample (e.g., Amazon Employee) and low sample (others) datasets. 

\subsection{Examining The Performance Trajectory of Iterative Explorations}
\begin{figure}[h]
    \centering
    \subfigure[German Credit]{
    \begin{minipage}[ht]{0.4\linewidth}
    \centering
    \includegraphics[width=1.3in]{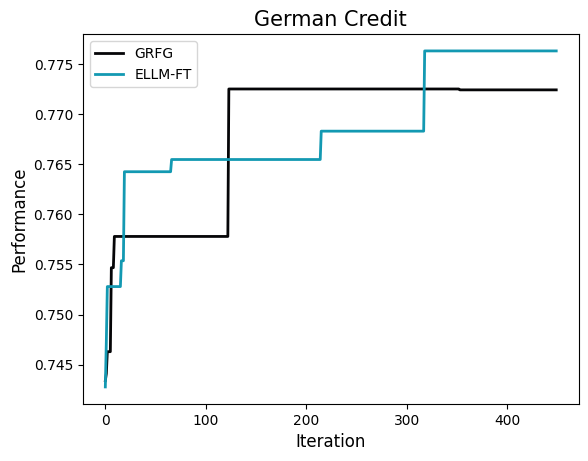}
    \label{exp:ger}
    \end{minipage}
    }
    \centering
    \subfigure[Openml\_586]{
    \begin{minipage}[ht]{0.4\linewidth}
    \centering
    \includegraphics[width=1.3in]{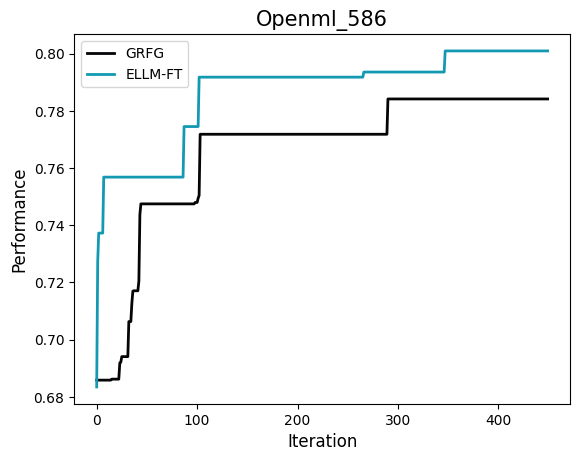}
    \label{exp:open}
    \end{minipage}
    }
    \caption{The performance trajectory of the proposed method compared to GRFG on two datasets.}
    \vspace{-0.4cm}
    \label{exp:trajectory}
\end{figure}

This experiment aims to answer: \textit{Can our evolutionary LLM method effectively and efficiently learn and generate an optimal sequence of feature transformation operations?}
We benchmarked our method, ELLM-FT, against GRFG,  a state-of-the-art fast reinforcement feature generation method.  
Figure \ref{exp:trajectory} shows the comparative performance trajectories of ELLM-FT and GRFG on the German Credit and Openml-586 datasets. 
Our findings are:
Our key findings are as follows: 
1. Early Iterations (First 50 Iterations): ELLM-FT demonstrates a sharper optimization trajectory, quickly generating feature transformation sequences with higher accuracy. This early advantage is likely due to the robust database constructed by our RL data collector, which offers a strong initial foundation for the search process. 
2. Long-Term Performance (Overextended iterations): ELLM-FT consistently evolves superior feature transformation sequences, showing more performance improvements compared to GRFG. 
This indicates that ELLM-FT not only leverages general feature knowledge but also excels in perceiving optimization directions even with few-shot samples. 
3. Overall Efficiency: ELLM-FT achieves high-performance results more rapidly than GRFG, showcasing its ability to optimize feature transformation sequences efficiently. Moreover, the continuous optimization observed in ELLM-FT’s trajectory suggests that it has the potential to uncover even more optimal results with further iterations. 

\subsection{A Study of Prompt Design}
\begin{figure}[h]
    \centering
    \subfigure[Downstream task accuracy]{
    \begin{minipage}[ht]{0.47\linewidth}
    \centering
    \includegraphics[width=1.3in]{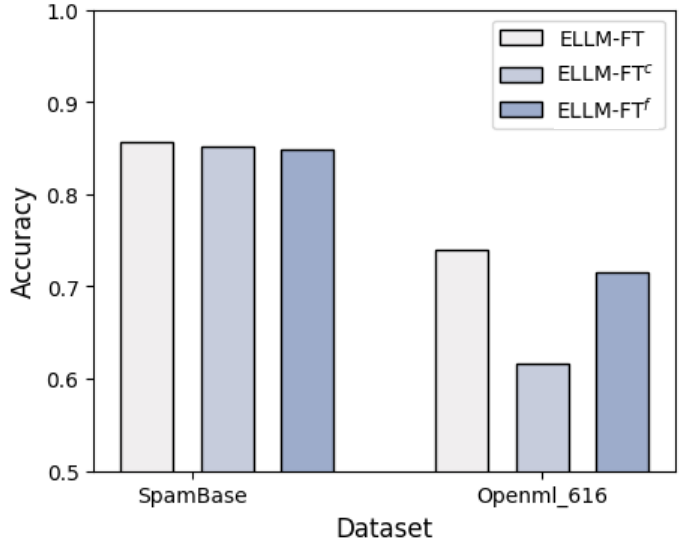}
    \label{exp:acc}
    \end{minipage}
    }
    \centering
    \subfigure[Valid sample numbers]{
    \begin{minipage}[ht]{0.47\linewidth}
    \centering
    \includegraphics[width=1.3in]{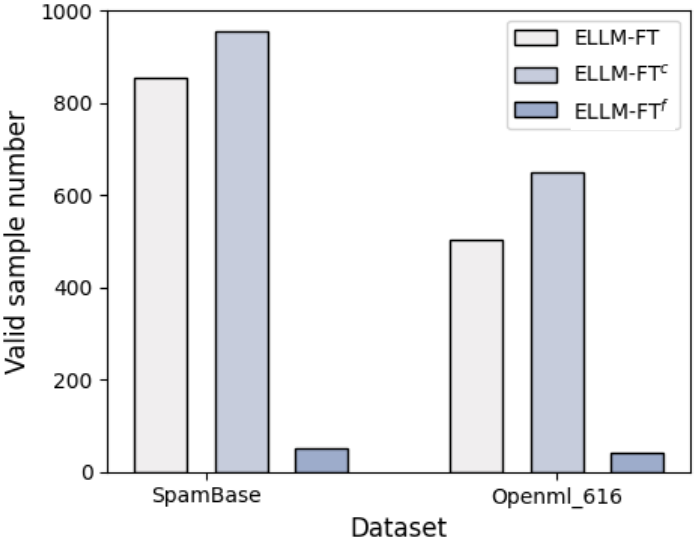}
    \label{exp:num}
    \end{minipage}
    }
    \caption{Results of the proposed method using three different prompts. We compared (a) downstream task accuracy across two datasets, and (b) valid sample numbers.}
    \vspace{-0.4cm}
\end{figure}

This experiment aims to address the question: \textit{Why is this particular prompt design necessary?} Our method leverages the Funsearch and randomness to balance the diversity (over populations) and quality (measured by performance) of  feature transformation sequences in prompts presented to LLM. We developed two variants of our model: 1) ELLM-FT\(^f\): This variant employs the prompt design inspired by Funsearch~\cite{romera2024mathematical}, where the top-$M$ samples in each iteration are selected based on their performance. 2) ELLM-FT\(^c\): This variant uses the prompt strategy from LMX~\cite{meyerson2023language}, which selects $M$ random samples in each iteration. 
1) Figure \ref{exp:acc} shows that our proposed prompt design enables the model to achieve superior downstream task performance on both datasets. 
2) Figure \ref{exp:num} presents the number of valid samples generated by each model after the same number of iterations. Although ELLM-FT\(^c\) generates the highest number of valid samples, its reliance on random and unordered samples hinders the model's ability to identify optimization directions, especially on noisier datasets, leading to suboptimal performance. The ELLM-FT\(^f\) variant assists the LLM in feature optimization by focusing on the top-$M$ samples. However, this approach can lead to under-utilization of other potentially useful data. As iterations progress, the model struggles to generate diverse and valid samples from repetitive inputs, resulting in a significantly lower number of valid samples compared to the other two models. Consequently, its long-term effectiveness is limited. 3) When considering both the number of valid samples and downstream task accuracy, ELLM-FT demonstrates the best overall performance. This highlights the efficacy of the proposed prompt design, as it balances the need for diversity with the ability to discern optimization directions effectively.

\subsection{Examining the Impact of RL Data Collector}

\begin{figure}[h]
    \centering
    \includegraphics[width=0.7\linewidth]{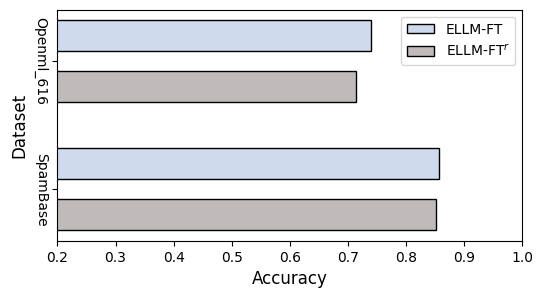}
    \caption{Results of the proposed method using the RL data collector and random selection.}

    \label{exp:rl}
    \vspace{-0.4cm}
\end{figure}

This experiment seeks to address the question: \textit{Is the database constructed by the RL data collector effective?} To assess this, we introduce a model variant, ELLM-FT$^r$, specifically designed to evaluate the impact of the RL data collector on feature transformation. Figure \ref{exp:rl} indicates that ELLM-FT consistently outperforms ELLM-FT$^r$ across both datasets.
This outcome suggests that the database constructed by the RL data collector inherently captures the optimization trajectories guided by reinforcement learning. This enables the model to generate more effective and optimized samples. Additionally, the RL data collector gathers a diverse and high-quality dataset, which further enhances the model's overall performance.

\subsection{Robustness Check}
This experiment aims to answer: \textit{Is ELLM-FT robust to different downstream models?} We replace the downstream models with K-Nearest Neighborhood (KNN), Decision Tree (DT), Support Vector Machine (SVM) and Ridge to study the variance of performance on the SVMGuide3 dataset.  Table \ref{tab:robust} shows that ELLM-FT consistently achieves the best performance regardless of the downstream models compared with baselines. The potential driver is that ELLM-FT can customize optimization strategies for various downstream models. Based on the feedback of the downstream models, it continually optimizes the feature sequence, thereby evolving the optimal individual. Thus, ELLM-FT demonstrates robustness to the downstream models. More experiments can be found in the Appendix.

\begin{table}[h]
    \centering
    \caption{The robustness check with different downstream models on SVMGuide3.}
    \label{tab:robust}
    \resizebox{0.8\linewidth}{!}{
    \begin{tabular}{cccccccccccc}
   \toprule\toprule
     & KNN & DT & SVM & Ridge & RF \\
     \midrule
     RDG & 0.717 & 0.780 & 0.780 & 0.773 & 0.789 \\
     ERG & 0.631 & 0.731 & 0.638 & 0.780 & 0.764 \\
     LDA & 0.699 & 0.680 & 0.719 & 0.719 & 0.732 \\
     AFT & 0.800 & 0.769 & 0.820 & 0.816 & 0.829 \\
     NFS & 0.717 & 0.760 & 0.801 & 0.781 & 0.831 \\
     TTG & 0.721 & 0.768 & 0.774 & 0.781 & 0.804 \\
     GRFG & 0.812 & 0.806 & 0.821 & 0.826 & 0.850 \\
     \midrule
     ELLM-FT & \textbf{0.823} & \textbf{0.810} & \textbf{0.830} & \textbf{0.843} & \textbf{0.856} \\
     
    \bottomrule\bottomrule
    \end{tabular}}
    \vspace{-0.4cm}
\end{table}


\section{Related work}
\subsection{Automated Feature Transformation}
AFT aims to reconstruct the feature space by automatically transforming the original features through mathematical operations. \cite{ying2024unsupervised,ying2023self,hu2024reinforcement} The prior literature can be categorized into three classifications:
1) expansion-reduction methods. These methods expand the feature space and filter out valuable features. DFS \cite{kanter2015deep} is first proposed to transform all original features and select significant ones. Cognito \cite{khurana2016cognito} performs feature transformation searches on a transformation tree and devises an incremental search strategy to efficiently explore valuable features. Furthermore, Autofeat \cite{horn2020autofeat} iteratively samples features based on beam search. 
2) evolution-evaluation methods. These methods are primarily grounded in genetic programming and reinforcement learning, iteratively navigating the decision process.. Binh et al. \cite{tran2016genetic} apply genetic programming to feature transformation. TransGraph \cite{khurana2018feature} uses Q-learning to decide on feature transformation. GRFE \cite{wang2022group} introduces three agents to collaborate in generating new feature transformations. 3) NAS-based methods. NAS \cite{zoph2016neural} regards the network architecture as a variable-length string. The use of reinforcement learning enables iterative exploration of network architecture. These methods can also be used in AFT. For example, NFS \cite{chen2019neural} employs an RNN-based controller to generate new feature transformations and train them through reinforcement learning. In this paper, we propose an evolutionary LLM framework for feature transformation. Our method distinguishes itself from prior literature in three main aspects: 1) \textit{diverse and high-quality search.} We maintain a multi-population database, enabling us to explore a wider range of potential populations at the search outset. Subsequently, through elimination and evolution, we conduct high-quality iterations; 2) \textit{implicit optimization direction.} Existing methods often lack an optimization direction or possess explicit ones. We introduce an implicitly optimized approach, encouraging LLMs to perceive potential optimization directions from few-shot prompts, facilitating flexible optimization; 3) \textit{generalization.} Our model demonstrates the ability to generalize across diverse domains, requiring only few-shot samples for generating new instances.

\subsection{LLM and Evolutionary Algorithms}
The Large Language Model (LLM) plays a significant role across various domains due to its formidable generative capabilities and understanding of natural language. However, LLM sometimes provides inaccurate responses due to insufficient external knowledge or memory biases, referred to as the hallucination of LLM \cite{shojaee2024llm,tonmoy2024comprehensive}. Various works have been proposed to address this issue \cite{lewis2020retrieval,gao2022rarr,varshney2023stitch, cheng2023uprise,jones2023teaching}, among which the perspective of self-refinement through
feedback and reasoning has garnered considerable attention \cite{madaan2024self, yang2024leandojo}. In this context, recent research has revealed that the integration of LLM with evolutionary algorithms not only alleviates the hallucination of LLM but also improves efficiency \cite{suzuki2024evolutionary}. LLM harnesses prior knowledge to continually engage in adaptive crossover and mutation, yielding significant strides in fields such as neural architecture search \cite{chen2024evoprompting}, symbolic regression \cite{shojaee2024llm}, and mathematical discovery \cite{romera2024mathematical}. In this paper, we propose an evolutionary LLM framework for automated feature transformation. By leveraging LLM's capacity for understanding sequential knowledge, the proposed method captures correlations between features and operations, thereby facilitating implicit optimization.
\section{Conclusion}
We introduce an evolutionary LLM-based feature transformation model. Our approach achieves automated feature transformation through two steps: 1) automatic construction of a multi-population database via RL data collectors; 2) feature transformation operation sequence search by few-shot prompting LLM. 
We optimize feature transformation using LLM based on the pairing of general feature knowledge and task-specific knowledge. Within the framework of evolutionary algorithms, we continuously iterate through evolutionary processes and culling operations. By maintaining multiple population databases, we achieve diverse and high-quality searches. Through extensive experimentation, we substantiate the effectiveness and practical impact of ELLM-FT.
\section*{Acknowledgment}
This research was partially supported by the National Science Foundation (NSF) via the grant numbers: 2426340, 2416727, 2421864, 2421865, 2421803, 2416728, and the National academy of engineering Grainger Foundation Frontiers of Engineering Grants.

%

\bibliography{ref}

\end{document}